\documentclass[letterpaper, 10 pt, conference]{ieeeconf}  % Comment this line out if you need a4paper
\usepackage{cite}
\usepackage{hyperref}
\usepackage{balance}
\usepackage{amsmath,amssymb,amsfonts}
\usepackage{pifont}  % for 圆圈序号
\usepackage{caption}  % 调整表格 caption
\usepackage{graphicx}
\usepackage{float}
\usepackage{bm}
\usepackage{multirow}
\usepackage{multicol}
\usepackage{array}
\usepackage{booktabs}

\makeatletter

\newcommand{\Rmnum}[1]{\expandafter\@slowromancap\romannumeral #1@}
\makeatother

\IEEEoverridecommandlockouts                              % This command is only needed if 
\title{\LARGE \bf
Intention-based and Risk-Aware Trajectory Prediction for Autonomous Driving in Complex Traffic Scenarios
}

\author{Wen Wei$^{1}$ and Jiankun Wang$^{1,2}$ \emph{Member, IEEE}
\thanks{This work is partially supported by Shenzhen Science and Technology Program under Grant RCBS20221008093305007, 20231115141459001, Young Elite Scientists Sponsorship Program by CAST under Grant 2023QNRC001, High level of special funds (G03034K003) from Southern University of Science and Technology, Shenzhen, China. (Corresponding author: Jiankun Wang)}
\thanks{$^{1}$Shenzhen Key Laboratory of Robotics Perception and Intelligence, Department of Electronic and Electrical Engineering, Southern University of Science and Technology, Shenzhen, China.
        {\tt\small wangjk@sustech.edu.cn}}%
\thanks{$^{2}$Jiaxing Research Institute, Southern University of Science and Technology, Jiaxing, China.}%
}

\begin{document}

\maketitle
 \thispagestyle{empty}
\pagestyle{empty}

%%%%%%%%%%%%%%%%%%%%%%%%%%%%%%%%%%%%%%%%%%%%%%%%%%%%%%%%%%%%%%%%%%%%%%%%%%%%%%%%
\begin{abstract}
Accurately predicting the trajectory of surrounding vehicles is a critical challenge for autonomous vehicles. In complex traffic scenarios, there are two significant issues with the current autonomous driving system: the cognitive uncertainty of prediction and the lack of risk awareness, which limit the further development of autonomous driving. To address this challenge, we introduce a novel trajectory prediction model that incorporates insights and principles from driving behavior, ethical decision-making, and risk assessment. Based on joint prediction, our model consists of interaction, intention, and risk assessment modules. The dynamic variation of interaction between vehicles can be comprehensively captured at each timestamp in the interaction module. Based on interaction information, our model considers primary intentions for vehicles to enhance the diversity of trajectory generation. The optimization of predicted trajectories follows the advanced risk-aware decision-making principles. Experimental results are evaluated on the DeepAccident dataset; our approach shows its remarkable prediction performance on normal and accident scenarios and outperforms the state-of-the-art algorithms by at least 28.9\%  and 26.5\%, respectively. The proposed model improves the proficiency and adaptability of trajectory prediction in complex traffic scenarios. The code for the proposed model is available at https://sites.google.com/view/ir-prediction.
%----------------------------github链接--------------------------------------------
\end{abstract}

%%%%%%%%%%%%%%%%%%%%%%%%%%%%%%%%%%%%%%%%%%%%%%%%%%%%%%%%%%%%%%%%%%%%%%%%%%%%%%%%
\section{INTRODUCTION}
In complex traffic environments, accurately predicting the trajectories of surrounding vehicles, as human drivers do, remains a major challenge for autonomous vehicles (AV). Our focus is on addressing the issue of trajectory prediction for AV in these complex scenarios.

%边缘预测的问题：我们使用联合预测
A vehicle's behavior is influenced not only by its historical movements but also by the actions of surrounding vehicles. To address this issue, Tolstaya et al. \cite{c1} propose a model based on Conditional Marginal Prediction (CMP), which predicts the future trajectories of other vehicles based on the queried future trajectory of the AV. However, a key drawback of CMP is that the AV can only react passively to other vehicles' predicted behaviors, even in critical situations like merging, lane changes, or unprotected left turns. In these scenarios, it is crucial for the AV to actively coordinate with other agents rather than merely react to predictions. Therefore, our model adopts a joint prediction setting, simultaneously predicting the trajectories of multiple agents, effectively modeling future interactions between them.
% enabling more human-like decision-making by incorporating both the AV's intentions and the surrounding agents' behaviors into the prediction process.
\begin{figure}
  \centering
  \includegraphics[width=1\linewidth]{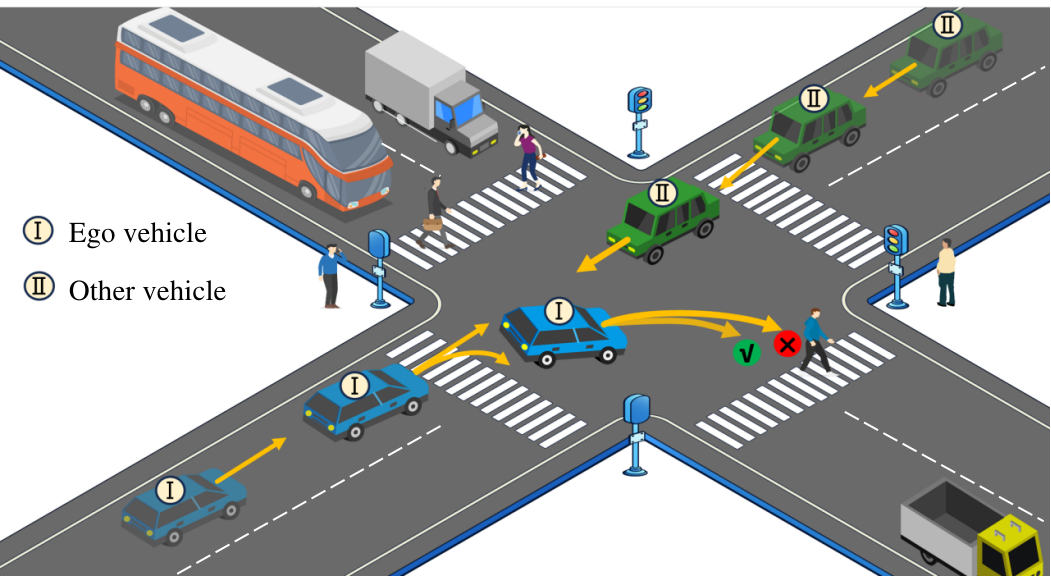}
  \caption{\Rmnum{1} represents the ego vehicle, \Rmnum{2} represents the other vehicle. This figure illustrates the process from trajectory generation to optimization in a multi-vehicle scenario. Our model generates various feasible vehicle trajectories, which are adjusted based on intention prediction. If the model detects that the AV's trajectory intersects with a high-risk area, such as a sidewalk, the trajectory is optimized to ensure both safety and adaptability.
%这张图展示了一个多车场景中从轨迹生成到优化调整的过程。
  }
\label{Intro}
\end{figure}

%不确定性的问题：我们采用意图识别
    Due to the inherent randomness and uncertainty in driver behavior, multiple reasonable trajectory choices may exist even in identical situations. In the face of such uncertainty, traditional prediction models often generate multiple possible trajectories \cite{c33,c2}, which increases decision-making complexity. However, intention recognition can effectively reduce this uncertainty by identifying the most likely driving behavior or trajectory in a given situation. Therefore, our model prioritizes trajectory prediction based on intention rather than considering all possible movement modes, making decisions more efficient and accurate.

    In complex mixed-traffic scenarios, especially those with a risk of collision, decision-making strategies must focus on accident avoidance. Existing approaches can be broadly categorized into classical methods \cite{c3,c4} and learning-based methods \cite{c5,c6,c7}. Classical methods, such as model predictive control (MPC) combined with potential field techniques \cite{c34}, typically rely on fixed parameters. While simple and efficient, learning-based methods often depend on manually set factors and parameters. In contrast, our approach incorporates risk assessment to evaluate the generated trajectories, enhancing decision-making in high-risk scenarios.

    Our model addresses the uncertainty problem by introducing intention prediction based on driving behavior, which constrains the generated trajectory. Additionally, by applying the principle of risk ethics, our model improves adaptability in complex scenarios, as shown in Fig.\ref{Intro}. The major contributions of this work are summarized as follows:%本研究重点探讨如何在复杂动态的交通环境中，结合预测模型的认知不确定性，提出一种安全的决策方法，以提高自动驾驶系统的安全性和可靠性。 

\begin{itemize}

\item We propose a novel intention feature module that enhances trajectory prediction by focusing on intention-related trajectories rather than considering all possible movement modes, thereby improving the accuracy and efficiency of the autonomous driving (AD) system.

\item We introduce a trajectory optimization module that accounts for potential risks, allowing vehicles to make more adaptable and safer decisions.

\item In both normal and accident scenarios, our model outperforms state-of-the-art (SOTA) baselines by at least 28.9\% and 26.5\%, respectively, demonstrating its effectiveness and adaptability in complex traffic environments.

\end{itemize}

\section{RELATED WORK}
\textbf{Multi-agent Trajectory Prediction.}
    Multi-agent joint prediction methods aim to generate consistent future trajectories for all agents of interest, thereby capturing interactions more effectively. The core of this approach lies in utilizing scene context and agent interaction information \cite{c9,c10,c11}. Early works often represent the scene context using bird's-eye view images, leveraging convolutional neural networks (CNNs) for feature fusion \cite{c12,c13}. However, these approaches may suffer from information loss and a limited receptive field.

    Recently, vectorized scene encoding schemes become popularity. In these methods, each scene context is represented as a vector and processed using techniques like graph convolutional networks (GCNs) \cite{c14,c15}, which help avoid the limitations of earlier methods. Agent interactions and their relationships with elements on the local map can be effectively modeled using Transformer modules \cite{c16,c18}. By modeling dynamic interaction dependencies across different timestamps, these methods can generate multiple realistic trajectories that better align with real-world scenarios \cite{c17}.

% 多智能体联合预测方法旨在以一致的方式为所有感兴趣的智能体生成未来轨迹，从而更好地捕捉智能体之间的交互。多智能体预测方法的核心在于利用场景上下文和智能体间的交互信息[1,2,3,4]。早期的工作通常将场景上下文化为鸟瞰图像，然后使用卷积神经网络（CNN）进行特征融合[1][4]。然而，这种方法可能会引入信息丢失并限制感受野。近年来，基于向量化的场景编码方案变得越来越流行，每个场景上下文被表示为向量，通过图卷积网络（GCN) [7]或Transformer[8,9,10,11]等技术进行处理，避免了早期方法的不足。通过利用Transformer模块可以对代理之间的交互及其与本地地图上不同元素的关系进行建模[12]。通过对不同时间戳之间的动态交互依赖性进行建模[13]，可以生成符合实际情况的多种合理轨迹。

\textbf{Risk-based Decision-Making.}
In AD systems, traditional motion planning methods often focus on optimizing paths or trajectories for optimal control and performance\cite{wang1,wang2}. Related work includes the use of extended Kalman filter methods to handle and propagate uncertainty in the future positions of surrounding vehicles \cite{c29}. These methods typically rely on fixed parameters, which can result in overly conservative plans. Similar issues arise in reachable set analysis methods \cite{c30,c31}.

These methods generally assume that the traffic environment is static or that obstacles are accurately predicted, assumptions that may not hold in dynamic and complex road conditions. To address this, risk-based planning approaches have been developed to account for the inherent uncertainties of road traffic, enhancing system safety by quantifying and minimizing potential risks.

Recent risk-aware architectures for AV incorporate uncertainties in predictive models, such as perception, intention detection, and control \cite{c19}. Risk measurement can be further extended by integrating the severity of potential collisions based on these uncertainties \cite{c20,c21}. While most risk-aware trajectory planning approaches focus on minimizing the risk or uncertainty for the AV itself, ethical considerations in risk assessment are increasingly important and must also account for the risks posed to other traffic participants\cite{c22}.
% 在自动驾驶系统中，传统的运动规划方法往往侧重于优化路径或轨迹，以实现最优的控制和性能。相关工作有使用扩展卡尔曼滤波器方法处理和传播周围车辆未来位置的不确定性[33]。进一步，采用MPC结合势场技术来处理预测的不确定性。然而，这些方法通常使用固定参数，这可能导致计划过于保守。类似的问题也出现在可达集分析方法中[34]、[35]。这些方法通常假设交通环境是静态的或对障碍物有准确预测，然而在动态和复杂的道路条件下，这种假设可能不再有效。为了应对这一挑战，基于风险的规划方法应运而生，它承认道路交通的固有风险，通过量化和最小化潜在风险来提升系统的安全性。用于AV的进一步风险感知架构纳入了有关预测模型认知的不确定性，例如感知、意图检测或控制[21]。基于不确定性，可以通过整合潜在碰撞的严重程度来扩展风险测量[22]、[23]。大多数这些风险意识轨迹规划方法旨在最大限度地减少自我自动驾驶的风险或不确定性。然而，从伦理角度对自动驾驶汽车风险评估的要求更为深远，需要纳入其他交通参与者的风险[5]。

% 另一种方法是使用概率预测技术，其中决策系统根据概率预测结果来规划运动。这些方法主要包括Branch MPC方法[36]和Stochastic MPC[37]。然而，这些方法也依赖于预先训练的预测模型，但没有考虑预测模型的认知不确定性。Prediction Failure Risk-Aware Decision-Making for Autonomous Vehicles on Signalized Intersections

\section{PRELIMINARIES}
% The neural network predictor receives two types of input data: historical states and scene context. The historical state refers to the dynamic features of each agent over the past $H$ time steps, including position, heading angle, and bounding box size. The AV, identified as $A_0$, and multiple surrounding traffic participants denoted as $A_1, \ldots, A_n$ are stored in a fixed-size tensor, with missing agents filled with zeros.The scene context $M$ includes two types of map elements: lane lines and environmental semantic information. Lane lines are represented by a series of vectors, and the environmental semantic information includes details such as weather conditions and the time of day. For easier processing, the positional attributes of all agents and map elements are transformed into the local coordinate system of the AV.
%输入表示。神经网络预测器接收两种类型的输入数据：历史状态和场景上下文。对于每个智能体，其历史状态是过去时间步 H 的一系列动态特征，包括其位置、航向角和边界框大小。我们考虑 AV 周围最近的 N 个智能体，其观测数据存储在固定大小的张量中，缺失的智能体用零填充。对于场景上下文，我们考虑两种类型的地图元素，即车道线和环境语义信息。车道线由一系列向量表示，环境语义信息中包括对当前场景的描述：天气类别和time of the day。所有代理和地图元素的位置属性都会转换为 AV 的本地坐标系。

We assume that the driving scenario can be described as a continuous space-discrete time system involving the autonomous vehicle (AV), denoted as $A_0$, and other agents, labeled as $A_1$ to $A_N$. The states of these agents are influenced by the scenario context, $M$. Given the historical states $s$ of all agents over the previous $H$ time steps, we define:
\begin{equation}
 X = \{s^{0}_{-H:0},s^{1}_{-H:0},s^{2}_{-H:0},\ldots, s^{N}_{-H:0}\}
\end{equation}
where $s$ includes its position, yaw, vehicle type, lateral and longitudinal behavior, and scene context. 

We denote the set of $K$ possible future trajectories for all agents as:
\begin{equation}
 Y_k = \{\hat{s}_{1:T}^{0}, \hat{s}_{1:T}^{1}, \ldots, \hat{s}_{1:T}^{N} \},  k = 1, \ldots, K
\end{equation}
where $ Y_k$ represents the set of predicted states for agent $ i$ over time steps $ 1$ to $ T$, where each predicted trajectory is associated with a probability $ \{p_k \mid k = 1, \ldots, K\}$.
Finally, we consider the AV’s initial trajectory $\hat{s}_0^{1:T}$, the predictions of other agents, and the defined cost function to optimize the future trajectories.

%我们假设驾驶场景可以被描述为一个包含AV及其周围多种交通参与者连续空间离散时间系统，其中AV被标识为$A_0$，其他智能体为$A_1$....$A_n$。这些智能体的状态受到场景上下文$M$的影响。给定前$H$个时间步的所有智能体的历史状态$X = \{s_{0:N-H:0}\}$和场景上下文$M$，我们的目标是预测接下来$T$个时间步中所有智能体的$K$个可能的联合未来轨迹 $\{Y_k \mid k = 1, \ldots, K\}$。这里，$Y_k = \{\hat{s}_{0:N}^{1:T}\}$表示智能体在未来时间步$t$的预测状态，而每个预测轨迹的概率为$\{p_k \mid k = 1, \ldots, K\}$。

%最终，为了优化AV的未来轨迹，我们需要考虑AV的初始轨迹$\hat{s}_0^{1:T}$、其他智能体的预测结果以及定义的成本函数。

\begin{figure*}
  \centering
  \includegraphics[width=1\textwidth]{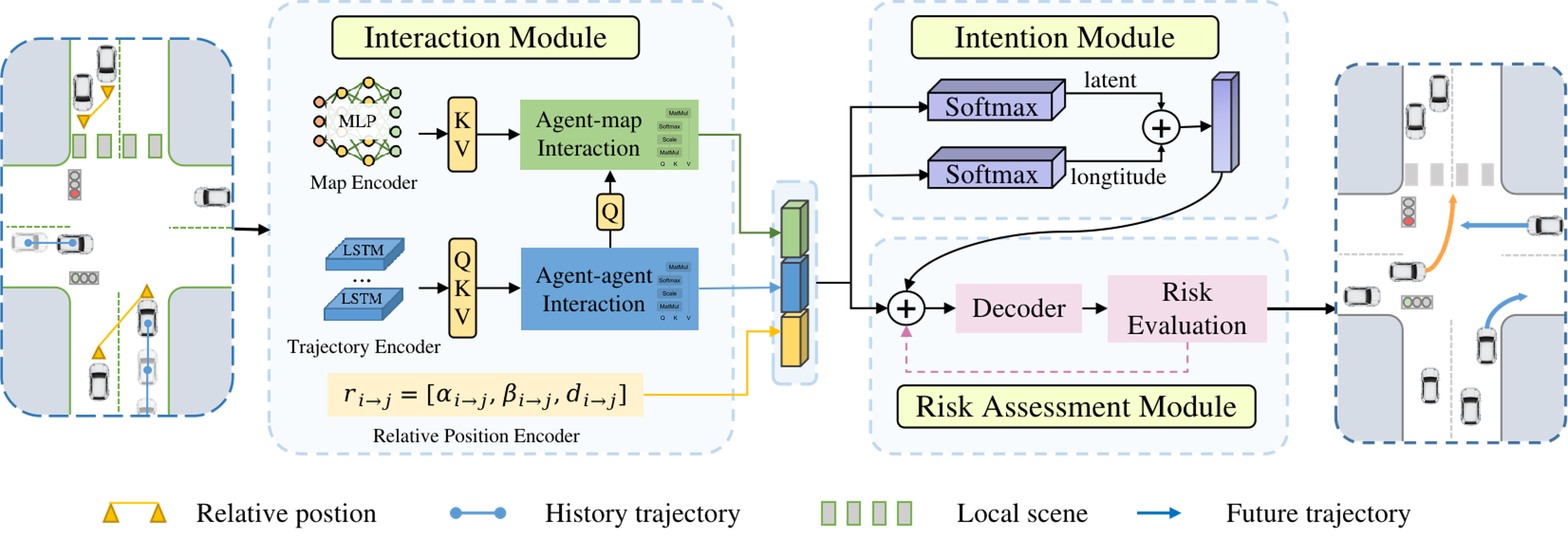}
  \caption{Illustration of the proposed model.}
  \label{model}
\end{figure*}

\section{PROPOSED MODEL}
Fig. \ref{model} shows the structure of our network. The interaction module (Sec.\ref{interactive}) processes historical trajectories and map information through the map encoder and trajectory encoder. The Relative Position Encoder further refines the positional relationships between agents. Subsequently, the interaction relationship between agents is modeled. In the Intention Module (Sec.\ref{maneuver}), the longitudinal and lateral intention probabilities are calculated and fused to obtain the agent's motion intention. Finally, the scene risk value is calculated through the Risk Assessment Module (Sec.\ref{risk}) and used to guide the trajectory optimization. By collaborating with different modules, the entire system can effectively predict and evaluate the risks in multi-vehicle interactions to ensure the proficiency and adaptability of driving in complex scenarios.
%图2展示了我们的网络结构。交互模块通过Local Scene Encoder和History Trajectory Encoder 处理历史轨迹和地图信息。Relative Position Encoder进一步处理车辆之间的位置关系。随后，对车辆间的交互关系进行建模。在Maneuver Module模块，计算纵向和横向的意图概率，并将其融合得到车辆的运动意图。最后，通过Risk Evaluation 计算场景风险值，并以此来指导车辆的轨迹优化。整个系统通过不同模块的协作，能够有效预测并评估多车交互中的风险，确保车辆安全行驶。

\subsection{Interaction Module} \label{interactive}
For each agent, a local scene context is constructed by gathering potential interactive scene contexts within a specified range. We compute the geometric properties of each agent and obtain the relative position embedding to help the model understand the positional changes of agents over time.For elements with absolute spatial-temporal positions $ ({d}_i, {v}_i, t)$ and $ ({d}_j, {v}_j, s)$, the relative position between element $ i$ and element $ j$ is descried by using three quantities: heading difference $ \alpha_{i \rightarrow j}$, relative bearing angle $ \beta_{i \rightarrow j}$, and distance $ {d}_{i \rightarrow j}$. To enhance numerical stability, angles are represented using sine and cosine values. We represent the heading difference $ \alpha_{i \rightarrow j}$ as $ \sin(\alpha_{i \rightarrow j}) = \frac{{v}_i \times {v}_j}{\|f{v}_i\|\|{v}_j\|}$, 
$ \cos(\alpha_{i \rightarrow j}) = \frac{{v}_i \cdot {v}_j}{\|f{v}_i\|\|{v}_j\|}$, 
and the relative bearing angle $ \beta_{i \rightarrow j}$
% (the angle between displacement vector $\rm f{d}_{i \rightarrow j}$ and heading vector ${v}_j$)
as $ \sin(\beta_{i \rightarrow j}) = \frac{{d}_{i \rightarrow j} \times {v}_j}{\|{d}_{i \rightarrow j}\|\|{v}_j\|}$, 
$ \cos(\beta_{i \rightarrow j}) = \frac{{d}_{i \rightarrow j} \cdot {v}_j}{\|{d}_{i \rightarrow j}\|\|{v}_j\|}$. 
The relative spatial-temporal information becomes 
$ {r}_{i \rightarrow j} = [\sin(\alpha_{i \rightarrow j}), \cos(\alpha_{i \rightarrow j}), \sin(\beta_{i \rightarrow j}), \cos(\beta_{i \rightarrow j}), \|{d}_{i \rightarrow j}\|]$. We connect the agent’s geometric properties with semantic attributes (such as the agent's category) and obtain relative position embedding through a multi-layer perception (MLP).
%对于每个智能体，我们构建一个本地场景上下文，在一定范围内收集它可能产生交互的场景上下文。我们计算智能体的几何属性，得到相对位置嵌入，以帮助模型了解智能体随时间变化的相对位置。对于具有绝对时空位置 (di, vi, t) 的元素和具有 (dj, vj, s) 的元素，我们使用三个量来描述元素 i 和元素 j 之间的相对位姿：航向差 αi→j、相对方位角 βi→j 和距离 ∥di→j∥。为了增强数值稳定性，角度使用正弦和余弦值表示。我们将航向差 αi→j 表示为 sin(αi→j ) = vi × vj ∥vi∥∥vj∥ , cos(αi→j) = vi · vj ∥vi∥∥vj∥ ，以及相对方位角 βi→j （位移向量 di→j 与航向向量 vj 之间的角度）为 sin(βi→j ) = di→j × vj ∥di→j ∥∥vj ∥ , cos(βi→j ) = di→j · vj ∥di→j ∥∥vj ∥ 。相对空间信息成为5维向量 ri→j = [sin(αi→j), cos(αi→j), sin (βi→j)，cos(βi→j)，∥di→j∥]

 %我们将智能体的几何属性与语义属性（例如智能体的类别）连接起来，并通过MLP以获得相对位置入。
Inspired by \cite{c27}, we use the long short-term memory (LSTM) network as an encoder for history trajectories. The map encoder uses an MLP to encode map embedding. We use a two-layer self-attention Transformer encoder as the agent-agent interaction encoder, where the query, key, and value (Q, K, V) are the encoded agents' historical trajectory embedding. We use a Transformer encoder as the agent-map encoder where the interaction features of agents are query (Q), and we use the map embedding (including the sequence of encoded waypoints) as keys and values (K, V). This operation is performed multiple times to process all map vectors from the agents, resulting in a series of agent-map vector attention features. We build an interaction model centered on each agent for a future time frame. Each agent's interaction graph is independently constructed based on its characteristics and the surrounding environment, allowing for a more precise capture of its unique behavior patterns and interaction needs. %我们使用两层自注意力 Transformer 编码器作为代理-代理交互编码器，其中查询、键和值（Q、K、V）是编码代理的历史轨迹嵌入。我们利用智能体的交互特征作为查询（Q），并使用单个地图向量（编码路径点的序列）作为键和值（K，V）。该操作被多次调用以处理来自代理的所有地图向量，从而产生一系列代理-地图向量注意特征。我们依次以每个智能体为中心构建其未来一定时间内的交互模型，每个智能体的交互图是根据其自身特性和周围环境独立构建的，这样可以更精确地捕捉到其独特的行为模式和交互需求。

% 个性化的交互建模能够更好地反映每个智能体在不同场景中的动态变化。这种方法能更好地捕捉不同场景中的多样性和泛化能力，通过层次化建模表达不同层次的交互关系。
\subsection{Intention Module} \label{maneuver}
Due to the complexity and diversity of possible driving intentions, the actual trajectory of a vehicle in real-world traffic scenarios often remains uncertain. We categorize the primary driving intentions into lateral directions (left turn (LT), straight (ST), and right turn (RT)) and longitudinal directions (acceleration (ACC), constant speed (CON), and deceleration (DEC)). To address the uncertainty and variability in predictions, we introduce an intention module responsible for predicting the probability distribution of these driving intentions. Specifically, we use MLP layers to transform interaction features from historical data into future predictions, generating intention-specific embedding. To get the intention feature $ Z$, intention-specific embedding is linked with the predicted probabilities of intention classes $ l_a$ and $ l_o$ by the MLP with a softmax activation function, where $ l_a \in \{\text{LT}, \text{ST}, \text{RT}\}$ and $ l_o \in \{\text{ACC}, \text{DEC}, \text{CON}\}$.
% 由于可能的驾驶操作的复杂性和多样性，车辆在真实交通场景中的实际轨迹通常是不确定的。我们认为，车辆在在横向和纵向主要驾驶操作分别是左拐、直行、右拐，加速、均速、减速。为了考虑预测中的不确定性和可变性，我们引入了maneuver module，它负责预测车辆不同驾驶操作的概率分布。具体来说，使用 MLP 层将交互特征从历史数据转换为未来数据，产生特定机动的特征 Z。利用具有softmax激活函数的全连接层来将意图特定特征 Z 和机动类别 la 和 lo 的预测概率连接起来。
\begin{equation}
 Z = \text{softmax}(\text{MLP}(e^{la} \oplus e^{lo}, W_{lo}))
\end{equation}

The intention feature $Z$ is concatenated with the interaction features $I$. The combined representation is then fed into a decoding network. To predict the probability distribution of each future joint trajectory (all agents), we use max pooling to aggregate information from all agents and employ a MLP to decode these probabilities. Our approach evaluates various potential intentions that the vehicle might execute and quantifies the confidence level associated with each prediction. This is particularly useful for making informed decisions about anticipated intentions, as it enables the autonomous vehicle (AV) to account for the inherent uncertainty in the predictions. The trajectory optimization process will use the predictions with the highest probability as input, including the initial plan and predictions of other agents.

% 我们将交互特征I与机动预测概率concatenated起来，然后输入到解码网络中，为了预测每个未来的概率（所有智能体的联合轨迹），我们使用 maxpooling 来聚合来自所有智能体的信息，并使用 MLP 来解码概率。我们考虑了车辆可以执行的多种潜在操作,还量化了与每个预测相关的置信水平。这对于针对预期操作做出明智的决策特别有益，因为它允许自车考虑预测中固有的不确定性。后续的轨迹优化将以最高概率将未来（即初始计划和其他代理的预测）作为输入。

\subsection{Risk Assessment Module} \label{risk}
As mentioned above, the constructed model outputs prediction results without considering any information about the risk of the scenario. However, the real-world traffic environment is dynamic and full of uncertainty. Optimizing output results require additional information, especially in scenarios where training data is insufficient or unavailable, which usually refers to unsafe and high-risk situations. Therefore, quantifying the risk of a scenario is crucial for developing a reliable and trustworthy autonomous driving system. In order to solve the problem, we introduce a risk assessment module based on principles of risk ethics to optimize the trajectories.
%如上所述，构建的模型（1）输出确定性预测结果，而不提供任何有关场景风险的信息。然而，现实世界的交通环境是动态的且充满不确定性，对于训练数据不足或不可用的长尾驾驶场景，通常指不安全和高风险的情况，自车的规划需要更多的信息。因此，量化场景的风险对于建立诚实可信的自动驾驶系统至关重要。为了实现预测模块和规划模块的耦合，我们引入了基于风险伦理原则的风险感知模块来优化轨迹。

\subsubsection{risk estimation}
Based on the agent's heading angle and dimensions, we calculate the front and rear positions of each agent, and these positions with the agent's center position are used for collision detection. The collision probability is defined as following a multivariate normal distribution. By calculating the collision probabilities at the center, front, and rear positions and summing them, we determine the overall collision probability for the vehicle at that moment.
%基于车辆的航向角和尺寸，我们计算出每辆车的前端和后端位置，并将这些位置与车辆的中心位置一起用于碰撞检测。碰撞概率定义为符合多元正态分布，通过计算中心、前端、后端三个位置的碰撞概率并将其累加，得出车辆在该时刻的整体碰撞概率。

When considering factors that impact collisions, for objectivity, we only take into account factors such as the masses, velocities, and deflection angles of the colliding parties, which are not subject to human alteration. Based on the collision angle, the type of collision is classified into front, side, and rear. To simplify the collision calculation, we apply symmetrical treatment to the collision areas.
%考虑到影响碰撞的因素时，为了客观起见，我们只考虑碰撞双方的质量、速度、偏转角等不因人为而改变的因素。根据碰撞的角度，将其归类为前方、侧方或后方碰撞。为了简化碰撞模型，我们对碰撞区域进行了对称处理。
Our model distinguishes between protected (vehicles, trucks, etc.) and unprotected (pedestrians, cyclists, etc.) agents. The harm calculation equation introduced by \cite{c22} is:
\begin{equation}
 \Delta v_A = \frac{m_B}{m_A + m_B} \sqrt{v_A^2 + v_B^2 - 2 v_A v_B \cos \theta }
\end{equation}
\begin{equation}
 H = \frac{1}{1 + e^{-\left(\mu_0 + \mu_1 \cdot \Delta v + \mu_{\text{area}}\right)}}
\end{equation}
where $ m$ and $ v$ are the mass and velocity of the two agents, \(\theta \) is the collision angle, and \(\mu_0\), \(\mu_1\), and \(\mu_{\text{area}}\) are empirically determined coefficients.
%我们的模型区分受保护（车辆、卡车等）和不受保护（行人、骑自行车者等）的道路使用者。由[x]引入损害计算方程：其中 m 和 v 是两个道路使用者 A 和 B 的质量和速度，α 是碰撞角度，c0、c1 和 carea 是根据经验确定的系数。

The risk model aims to assign the estimated damage to each collision probability and then get the risk that varies with time within the planning horizon for each sampled trajectory. %使用伤害模型，我们将估计的伤害分配给每个碰撞概率。因此，我们知道每个采样轨迹的规划范围内随时间变化的风险。
For convenience in subsequent calculations, we aim to describe each possible trajectory with a single risk value, so the maximum risk at future time steps is chosen as the risk value for each trajectory:
%为了方便后续计算，我们希望用一个风险值来描述每一个可能的轨迹，因此我们选择未来时刻里最大风险作为每个轨迹的风险值。我们仅计算自车与其他车辆可能发生的碰撞的综合风险。
\begin{equation}
 R = \max_t(H \cdot P_t)
\end{equation}
Conditionally, only the aggregated risk of collisions is computed that the AV might have with other vehicles.

\subsubsection{cost function}
The cost function contains a variety of carefully designed risk costs that take into account different critical factors in driving decisions, including self-protection (safety), concern for vulnerable groups (care), and response to sudden high risks (Responsiveness). Below are the details of the calculation of each type of cost.
% 成本函数包含各种精心设计的风险成本，这些成本对驾驶决策的不同方面进行考虑，包括自我保护、关注弱势群体、关注突兀的高风险。下面给出了不同成本计算的详细信息。

The safety cost is calculated by taking into account the ego risk, obstacle risk, and boundary harm:
\begin{equation}
 \bm{ c_s = \frac{1}{2n}(\sum_{i=1}^{n} R_i + R_b)}
 \label{cs}
\end{equation}
where $R_i$ means the risk from ego vehicle, $R_j$ from other vehicles and $R_b$ means the risk from collision with road boundary. The risk of all detected agents in the scene is accumulated and normalized. According to the safety principle, the trajectory with the lowest overall risk must be chosen in the cost function. This principle makes the best decision for all road users overall.
%通过综合考虑自我保护风险、障碍物风险以及边界损害，来计算轨迹的总体风险成本。对场景中所有检测到的智能体的风险进行累积并进行归一化。根据贝叶斯原理，成本函数中必须选择总体风险最低的轨迹。这一原则确保自动驾驶汽车为所有道路使用者整体做出最佳决策。
 \begin{figure}[h]
  % \centering
  \includegraphics[width=1\linewidth]{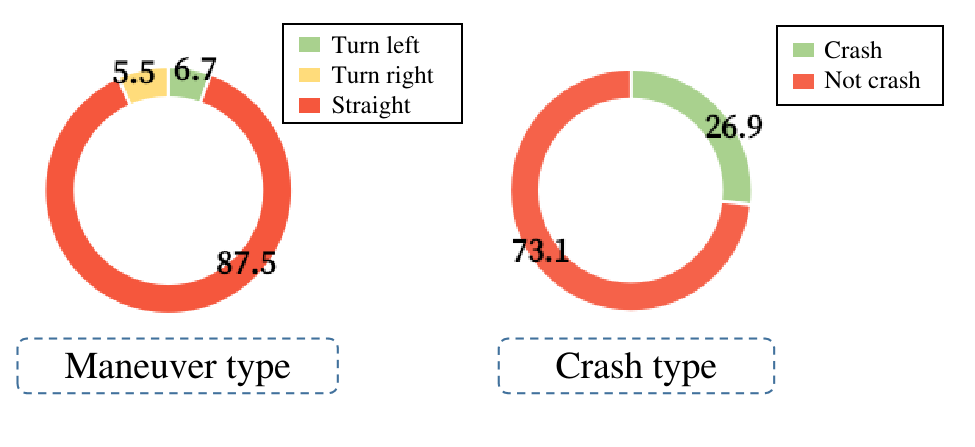}
  \caption{The proportion of different scenario subsets.}
  \label{dataset}
\end{figure}

To protect vulnerable groups (such as pedestrians and cyclists), we introduced the care cost:
\begin{equation}
\bm{ c_c = \frac{1}{n} \sum_{i=1}^{n}\sum_{j=1}^{n} \left| R_i - R_j \right|}
\end{equation}
that calculates the average difference between different risk values. The greater the difference between the risk of protected and unprotected groups, the higher the cost of care. This principle aims to avoid placing disproportionately high risks on vulnerable groups in the pursuit of lower safety costs.
%为了保护弱势群体（如行人和骑自行车者），我们引入了“关怀成本”，计算不同风险值之间的平均差异。如果受保护群体和非受保护群体风险之间的差异越大，关怀成本就越高。这种方式旨在避免在追求较低安全成本的过程中，对弱势群体造成不成比例的高风险。

In response to sudden high-risk situations, the responsiveness function evaluates the trajectory's risk cost based on the maximin principle, ensuring that the agent's performance remains acceptable even under the most adverse conditions (i.e. when the risk is high). 
\begin{equation}
 \bm{c_r = \max \left( \sum_{i=1}^{n} f(R_i) \right)}
\end{equation}

This principle evaluates the potential risk by calculating the maximin value between agents and uses a scaling factor to adjust the final risk cost. This means that when facing sudden high risks, the agent will prioritize scenarios that could cause the greatest harm, ensuring that the risk cost is effectively managed even in the worst-case scenario.
%应对突发高风险时，该函数根据极大极小原则评估轨迹的风险成本，以确保在最不利的情况下（即风险较高时）车辆的表现仍然能够接受。该原则通过计算自我保护和障碍物之间的极大极小值来评估潜在的风险，并使用缩放因子来调整最终的风险成本。这意味着，在面对突发高风险时，系统会优先考虑可能带来最大伤害的情况，以确保即使在最恶劣的条件下，风险成本也能被有效控制。

The final cost function is:
\begin{equation}
 L_{ risk} = \omega_s \cdot c_s +\omega_c \cdot c_c + \omega_r \cdot c_r
\end{equation}

Based on the final cost function, the agent can obtain the risk value related to the surrounding agents and further optimize the vehicle's trajectory. Among them, $\omega_s$ = $\omega_c$ = $\omega_r$ = 33.3, indicating that the model treats these three considerations equally.
%最终的cost function为：基于该函数生成的轨迹使自车能够得到与周围车辆相关的风险值，进一步优化自车轨迹。
\subsection{Model Training}
For the intention prediction, the cross-entropy loss is applied as follows:
\begin{equation}
 L_{ man} = -\sum_{la,lo} Q_{ gt} \cdot \log Q_{pre}
\end{equation}
where $ Q_{gt}$ and $ Q_{pre}$ represent the actual intention and predicted intention of the current training sample, respectively. For the trajectory prediction, the model selects the case closest to the real-world trajectory and then calculates the smoothed L1 loss.%其中，L_{gt}和L_{pre}分别表示当前训练样本的真实机动和预测机动。为了轨迹预测，我们选择最接近真实世界轨迹的情况，然后计算平滑的 L1 损失。其中 \hat{y}_i^{(k)}是与最接近真实情况的自我计划相关的预测分支，y_i\right 表示其他智能体的真实轨迹。
\begin{equation}
 L_{pre} = \min_{k} \sum_{i=1}^{N} \text{Smooth\_L1}\left(\hat{y}_i^{(k)}, y_i\right)
\end{equation}
where $ \hat{y}_i^{(k)} $is the predicted branch that is closest to the ground truth, and $ y_i$ represents the ground-truth trajectories.

\begin{figure*}[h!]
  % \centering
  \includegraphics[width=1\textwidth]{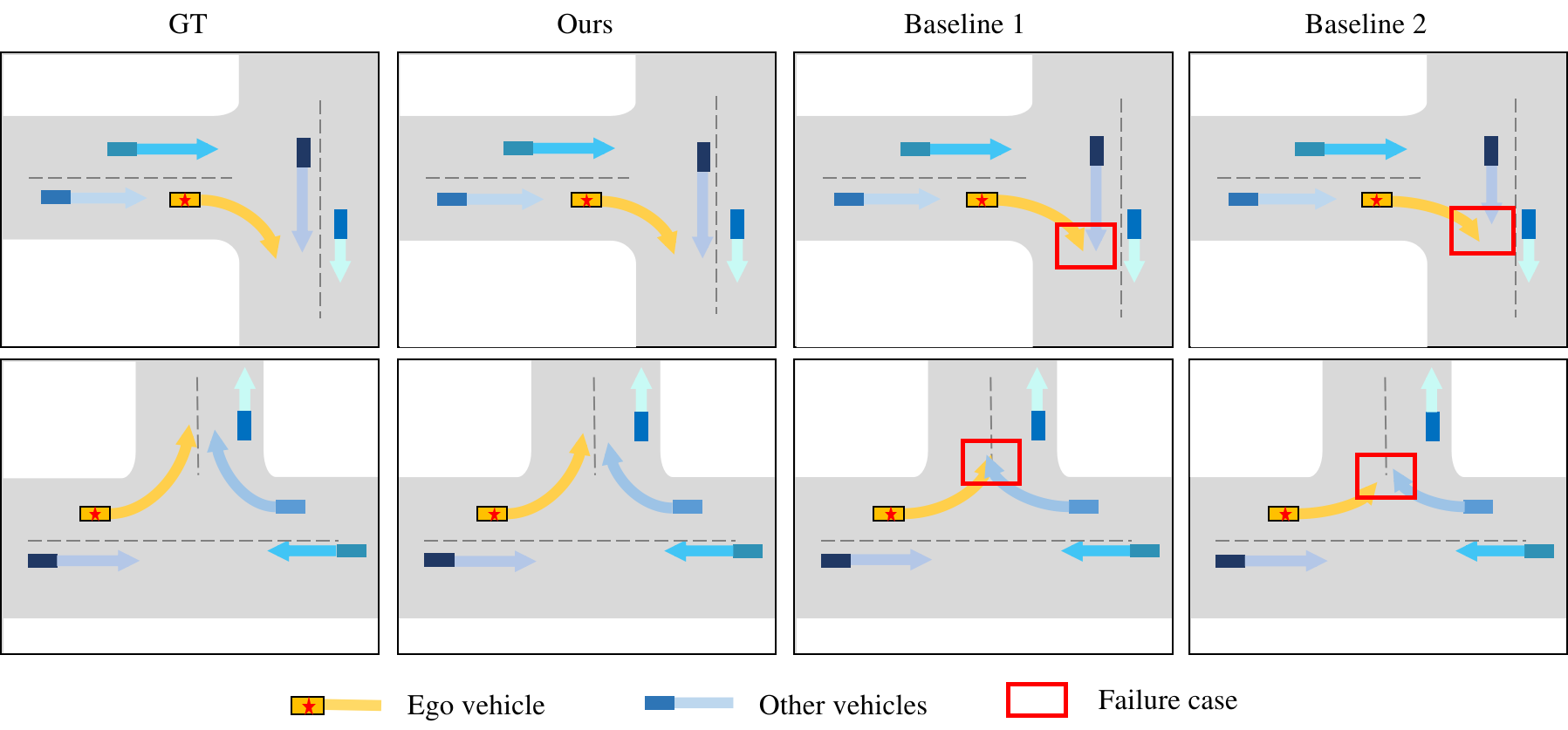}
  \caption{Illustration of prediction results in different traffic scenes. Top row: scenario exists for one right-turn operation. Bottom row: scenario exists collision probability. Compared with baselines, our model performs better in both scenarios, not only predicting accurately but also considering potential risks in the future.}
  \label{result show}
\end{figure*}

Similar to \cite{c28}, we set up a staged strategy. Particularly, in the first five epochs, we use the loss to evaluate the deviation between the observed and predicted trajectories. In subsequent epochs, the impact of risk is considered on trajectory generation. Therefore, the total loss is as follows:
%我们设置了一个分状态策略，特别是，在前五个epochs，我们利用损失来评估观察轨迹和预测轨迹的偏差，在后续的epochs，我们考虑风险对轨迹生成的影响。因此，the total loss 如下所示：
\begin{equation}
\begin{cases}
 L =L_{pre} + \tau \, L_{man}, & \text{Stage 1} \\
 L =L_{pre} + \tau \, L_{man} + (1-\tau \,)L_{risk}, & \text{Stage 2}
\end{cases}
\end{equation}

\section{EXPERIMENTS}
\subsection{Experiment Setup}

In this study, the DeepAccident \cite{c23} dataset is used for evaluation, which comprises a total of 285k annotated samples and 57k annotated V2X frames at a frequency of 10 Hz. Besides, we split the data with a ratio of 0.7, 0.15, and 0.15 for training, validation, and testing splits, resulting in 203k, 41k, and 41k samples, respectively. The dataset provides valuable insights into real traffic scenarios, allowing us to draw meaningful conclusions. In the experiment, we perform model training and inference on an NVIDIA GeForce RTX 4090 24GB GPU. To match this hardware platform, we implement our model on Ubuntu 18.04 and Pytorch 3.8 environment. In the process, we use a batch size of 32 and an Adam optimizer with a learning rate that starts from 2e-4 and a weight decay of 3e-4.

\begin{table}[h!]
\centering
\caption{Evaluation results for the proposed model and the baselines in the crash-based test set over a different horizon. Note: ADE is the evaluation metric, where lower values indicate better performance.}
\label{tab:my-table1}
\renewcommand{\arraystretch}{1.3} % Adjust this value for more or less spacing
\resizebox{\columnwidth}{!}{%
\begin{tabular}{cllllll}
\hline
\multirow{2}{*}{\textbf{Dataest}}                       & \multicolumn{1}{c}{\multirow{2}{*}{\textbf{Model}}} & \multicolumn{5}{c}{\textbf{Prediction Horizon(s)}}                      \\ \cline{3-7} 
                                               & \multicolumn{1}{c}{}                       & 1                        & 2                        & 3 & 4 & 5 \\ \hline
\multicolumn{1}{c|}{\multirow{6}{*}{NORMAL}}      & \multicolumn{1}{l|}{S-LSTM\cite{c24}}  & 1.32& 2.10& 3.15& 4.27  &5.54\\
% \multicolumn{1}{c|}{}                          & \multicolumn{1}{l|}{CS-LSTM}               &                          &                          &   &   &   \\
\multicolumn{1}{c|}{}                          & \multicolumn{1}{l|}{SGAN\cite{c26}}&  1.35& 2.09 &3.13&4.25 &5.53   \\
\multicolumn{1}{c|}{}                          & \multicolumn{1}{l|}{SGAN-P\cite{c26}}& 1.48 & 2.15 &3.32& 4.37  &5.63\\
\multicolumn{1}{c|}{}                          & \multicolumn{1}{l|}{Pishgu\cite{c32}}               &1.19   &    1.67    & 2.55  &  3.17  & 4.32  \\
\multicolumn{1}{c|}{}   & \multicolumn{1}{l|}{DIPP\cite{c27}}              &      0.33  & 0.89  &  1.70 & 2.69  & 3.29  \\

\multicolumn{1}{c|}{} & \multicolumn{1}{l|}{Ours} &\textbf{0.32} & \textbf{0.68} & \textbf{1.27}& \textbf{1.80} & \textbf{2.28} \\ \hline
\multicolumn{1}{c|}{\multirow{6}{*}{ACCIDENT}}   & \multicolumn{1}{l|}{S-LSTM\cite{c24}}  & 1.54 & 2.34 & 3.53& 4.88 & 6.40  \\
% \multicolumn{1}{c|}{}                          & \multicolumn{1}{l|}{CS-LSTM}               &                          &                          &   &   &   \\
\multicolumn{1}{c|}{}                          & \multicolumn{1}{l|}{SGAN\cite{c26}}  &1.53&2.33 &3.56& 4.87&6.31     \\
\multicolumn{1}{c|}{}                          & \multicolumn{1}{l|}{SGAN-P\cite{c26}} &1.64& 2.38 &3.67& 4.90  & 6.31\\
\multicolumn{1}{c|}{}                          & \multicolumn{1}{l|}{Pishgu\cite{c32}}               & 1.36  &    2.17  & 2.96   &  3.89  & 5.03  \\
\multicolumn{1}{c|}{}   & \multicolumn{1}{l|}{DIPP\cite{c27}}              &    0.92  &  1.27   &  2.54  & 4.26  & 5.89  \\

\multicolumn{1}{c|}{}   & \multicolumn{1}{l|}{Ours}     & \textbf{0.49}
&\textbf{0.83} &\textbf{1.88} &\textbf{3.02} &\textbf{4.69}\\ \hline
\multicolumn{1}{c|}{\multirow{6}{*}{ALL}} & \multicolumn{1}{l|}{S-LSTM\cite{c24}}        &   1.45 &  2.23 & 3.45 & 4.55 & 5.96\\
% \multicolumn{1}{c|}{}                          & \multicolumn{1}{l|}{CS-LSTM}               &                          &                          &   &   &   \\
\multicolumn{1}{c|}{}                          & \multicolumn{1}{l|}{SGAN\cite{c26}}  & 1.44 & 2.21 &3.44 &4.54  & 5.92\\
\multicolumn{1}{c|}{}                          & \multicolumn{1}{l|}{SGAN-P\cite{c26}}  & 1.56 &2.27 &3.50 & 4.63  &5.97\\
\multicolumn{1}{c|}{}                          & \multicolumn{1}{l|}{Pishgu\cite{c32}}               &1.18   &  2. 24    & 2.86   &  3.47  & 4.68 \\
\multicolumn{1}{c|}{}   & \multicolumn{1}{l|}{DIPP\cite{c27}}              &    0.72   & 1.01   &  2.06 & 3.53  & 4.02  \\

\multicolumn{1}{c|}{} & \multicolumn{1}{l|}{Ours} & \textbf{0.38} & \textbf{0.88} & \textbf{1.75} &\textbf{2.34} & \textbf{3.32}    \\ \hline
\end{tabular}%
 }
\end{table}
\begin{table}[h!]
\centering
\caption{Evaluation results for the proposed model and the baselines in the intention-based test set }
\label{tab:my-table2}
\renewcommand{\arraystretch}{1.3} % Adjust this value for more or less spacing
\resizebox{\columnwidth}{!}{%
\begin{tabular}{cllllll}
\hline
\multirow{2}{*}{\textbf{Dataest}}                       & \multicolumn{1}{c}{\multirow{2}{*}{\textbf{Model}}} & \multicolumn{5}{c}{\textbf{Prediction Horizon(s)}}                      \\ \cline{3-7} 
                                               & \multicolumn{1}{c}{}                       & 1                        & 2                        & 3 & 4 & 5 \\ \hline
\multicolumn{1}{c|}{\multirow{6}{*}{LEFT}}      & \multicolumn{1}{l|}{S-LSTM\cite{c24}}     & 2.44   &    4.36   & 5.95  & 7.97  & 10.77  \\
% \multicolumn{1}{c|}{}                          & \multicolumn{1}{l|}{CS-LSTM}               &                          &                          &   &   &   \\
\multicolumn{1}{c|}{}                          & \multicolumn{1}{l|}{SGAN\cite{c26}} &  2.45& 4.35  &5.94& 7.93 &10.75\\
\multicolumn{1}{c|}{}                          & \multicolumn{1}{l|}{SGAN-P\cite{c26}} & 2.65& 4.38 & 5.97& 7.98  &10.64\\
\multicolumn{1}{c|}{}                          & \multicolumn{1}{l|}{Pishgu\cite{c32}}               &2.52   &    4.10     & 5.63  &  8.12  & 10.39  \\
\multicolumn{1}{c|}{}   & \multicolumn{1}{l|}{DIPP\cite{c27}}   &                 1.05 & 1.38  & 2.66  & 4.35  & 6.23  \\

\multicolumn{1}{c|}{} & \multicolumn{1}{l|}{Ours} & \textbf{0.51}&\textbf{0.84}&\textbf{1.89}&\textbf{3.17}&\textbf{4.86} \\ \hline
\multicolumn{1}{c|}{\multirow{6}{*}{STRAIGHT}} & \multicolumn{1}{l|}{S-LSTM\cite{c24}}                & 1.32  &   2.09  & 3.06  & 4.24  & 5.32  \\
% \multicolumn{1}{c|}{}                          & \multicolumn{1}{l|}{CS-LSTM}               &                          &                          &   &   &   \\
\multicolumn{1}{c|}{}                          & \multicolumn{1}{l|}{SGAN\cite{c26}} & 1.33 &2.07 & 3.04& 4.23  &5.33\\
\multicolumn{1}{c|}{}                          & \multicolumn{1}{l|}{SGAN-P\cite{c26}} &  1.44& 2.13& 3.21& 4.37  & 5.42\\
\multicolumn{1}{c|}{}                          & \multicolumn{1}{l|}{Pishgu\cite{c32}}               &1.14   &    1.95     & 2.43   &  3.05  & 4.02  \\
\multicolumn{1}{c|}{}   & \multicolumn{1}{l|}{DIPP\cite{c27}}              &       0.54  & 0.99  & 1.65  & 2.79  & 3.40\\

\multicolumn{1}{c|}{}                          & \multicolumn{1}{l|}{Ours}   & \textbf{0.30}&\textbf{0.81}&\textbf{1.24}&\textbf{1.85}&\textbf{2.53}  \\ \hline
\multicolumn{1}{c|}{\multirow{6}{*}{RIGHT}}    & \multicolumn{1}{l|}{S-LSTM\cite{c24}}                &  2.45   &   4.35  & 5.99  &7.96   & 11.02  \\
% \multicolumn{1}{c|}{}                          & \multicolumn{1}{l|}{CS-LSTM}               &                          &                          &   &   &   \\
\multicolumn{1}{c|}{}                          & \multicolumn{1}{l|}{SGAN\cite{c26}} & 2.41 & 4.34 & 6.01& 7.94 &11.27  \\
\multicolumn{1}{c|}{}                          & \multicolumn{1}{l|}{SGAN-P\cite{c26}} & 2.58 & 4.36 & 5.92 & 8.00 & 10.74\\
\multicolumn{1}{c|}{}                          & \multicolumn{1}{l|}{Pishgu\cite{c32}}               &2.52   &    4.34     & 5.57   &  7.99  & 10.24  \\
\multicolumn{1}{c|}{}   & \multicolumn{1}{l|}{DIPP\cite{c27}}              &     0.99  & 1.36  & 2.67  & 4.32  & 6.02  \\

\multicolumn{1}{c|}{} & \multicolumn{1}{l|}{Ours} & \textbf{0.48} &\textbf{0.87} &\textbf{1.93}  &\textbf{3.15} &\textbf{4.75}   \\ \hline
\end{tabular}%
}
\end{table}

\subsection{Results}
To verify the effectiveness of the proposed model, we compare it with SOTA trajectory prediction models. These include well-known benchmarks such as S-LSTM\cite{c24}, SGAN\cite{c26}, Pishgu\cite{c32}, DIPP\cite{c27}. Note that, since SGAN inclues a refinement model with pooling, we introduce another baseline SGAN-P.

We use the Average Displacement Error (ADE) and Final Displacement Error (FDE) to evaluate the model's performance comprehensively. ADE refers to the average L2 distance between the ground truth and our predictions over all predicted time steps, while FDE measures the distance between the ground truth and our predictions at the final time step of the prediction period.

The results are displayed in Table \ref{tab:my-table1}. The driving scenarios are divided into collision and non-collision categories. Our model consistently outperforms the current SOTA baselines, with accuracy improvements ranging from 28.9\% to 61.1\% in normal scenarios, 26.5\% to 41.5\% in accident scenarios, and 27.3\% to 50.8\% overall. The results in accident scenarios are notable, given the high complexity and unpredictability of such scenarios. The improvement underscores the critical importance of incorporating driving behavior and risk perception into trajectory prediction. By effectively modeling the interactions between vehicles and assessing potential risks, our model can better navigate and predict outcomes in high-risk situations. In normal scenarios, our method initially performs similarly to DIPP but shows better long-term prediction capability. This enhanced long-term performance highlights the model's ability to maintain accuracy over extended prediction horizons, which is crucial for effective autonomous driving.

Additionally, we further categorize the data based on different vehicle intentions, including straight, left turn, and right turn, allowing us to conduct a detailed assessment of our model's capabilities in various traffic behaviors, as shown in Table \ref{tab:my-table2}. Specifically, in the straight driving test subset, our model achieves a significantly lower ADE value compared to the SOTA baseline, demonstrating the effectiveness of our approach in improving prediction accuracy. Furthermore, our model shows remarkable improvements in the left turn and right turn test subsets, highlighting its robustness and effectiveness in accurately predicting future vehicle trajectories across various driving scenarios and intentions. Overall, our findings confirm the capability and efficiency of our model in predicting vehicle trajectories.

%在这里，我们的模型始终超越当前的 SOTA 基线，Normal场景上的准确率提高了 28.9\% - 61.1\%，Accident场景上的准确率提高了 26.5\%-41.5\%， 总体场景提高了27.3\%-50.8\%。这些改进强调了考虑机动行为和风险感知意识的重要性，特别是在复杂的交通场景中。

%我们还对基于机动的测试集进行了测试，如表 2 所示。具体来说，在直行测试子集中，我们的模型实现了比 SOTA 基线显着更低的 ADE 值，说明我们的方法有效提高了模型的预测精度。此外，我们的模型在左拐和右拐测试子集中显示出显着的改进，突出了其在各种驾驶场景和操作中准确预测未来车辆轨迹的鲁棒性和有效性。总的来说，我们的研究结果证实了我们的模型在预测自动驾驶车辆轨迹方面的能力和效率。

\subsection{Ablation Study}
Table 3 shows the analysis of the four key components: interaction module, intention module, and risk assessment module. We tested five models: Model A (\textbf{without Interaction Module}), Model B (\textbf{without Intention Module}), Model C (\textbf{without Risk Assessment Module}), and Model D (\textbf{all model components}).% 表 3 展示了对四个关键组件的分析：交互模块、机动模块和风险感知模块。我们测试了五个模型：模型a（不包括交互模块））、模型b（不包括机动模块、模型c（不包括风险感知模块）和模型d（所有模型成分）

\begin{table}[h]
\centering
\caption{Evaluation results of ablation models}
\label{ab-study}
\renewcommand{\arraystretch}{1.4} % Adjust this value for more or less spacing
\resizebox{\columnwidth}{!}{%
\begin{tabular}{ccccc}
\hline
\textbf{Model} & \textbf{Normal ADE} & \textbf{Normal FDE} & \textbf{Accident ADE} & \textbf{Accident FDE} \\ \hline
A   & \text{1.60}   & 2.82       & \text{1.80}     & 3.88         \\
B     & 1.29      & 2.41      & \text{1.40}      & \text{2.62}          \\
C             & \text{1.26}    & \text{2.31}    & 1.46   & 2.65    \\
D      & \textbf{1.14} & \textbf{2.28}  & \textbf{1.39}  & \textbf{2.58} \\ \hline
\end{tabular}%
}
\end{table}

When evaluating the normal and accident scenarios, Models A, B, and C exhibited inferior performance compared to the comprehensive Model D. The performance of Model D highlights the significant impact of integrating multiple modules. The integration of the interaction module significantly improved performance, underling the importance of modeling interactive behaviors to enhance prediction accuracy. The intention module further enhances performance by effectively capturing driving intentions in various scenarios. Notably, Model C performs worse in accident scenarios compared to other models. This indicates that risk awareness is particularly critical in such high-risk situations. The performance of Model C underscores the importance of incorporating risk assessment into trajectory prediction systems. Without a robust mechanism for evaluating and responding to potential risks, models may fail to handle accident-prone scenarios effectively.

% Notably, model C performs worse in the Accident scenario, indicating that risk awareness is particularly crucial in accident-prone situations.
%在针对Normal和Accident场景进行评估时，与综合模型D相比，其他版本（A-C）均表现较差。交互模块的集成显着提高了性能，强调了对交互行为进行建模在提高预测准确性方面的重要性。机动模块在各种场景都能够进一步灵敏捕获动态车辆交互，提高了性能。值得注意的是，模型C在Accident场景中表现较差，这表明风险意识在事故场景中尤为重要。

\section{CONCLUSIONS}
Predicting the trajectories of surrounding vehicles in complex environments is necessary for AD system. To overcome this challenge, we proposed a joint prediction-based model consisting of three parts: an interaction module, a intention module, and a risk assessment module. Our model maintains high accuracy in the normal scenario and demonstrates the potential to cope with challenging or unusual situations in the accident scenario. Through an ablation study, the importance of each module is validated, and the need to incorporate the traffic behavior principles is emphasized. Overall, the performance of our approach verifies its proficiency and adaptability. In the future, we plan to study the planning module corresponding to the proposed prediction module and build a fully functional AD system.
%在复杂环境下预测周围车辆的轨迹是实现完整自动驾驶系统过程中必经之路。为了应对这一挑战，我们提出了一种基于联合预测的模型，由三个部分组成：交互模块，机动模块，risk-aware 模块。我们的模型在noraml场景中保持了应有的高精度效果，在accident场景下，展示了应对挑战性或不寻常的情况的潜力，综合来看，整体表现验证了我们方法的有效性，和模型的鲁棒性。


%%%%%%%%%%%%%%%%%%%%%%%%%%%%%%%%%%%%%%%%%%%%%%%%%%%%%%%%%%%%%%%%%%%%%%%%%%%%%%%%

%%%%%%%%%%%%%%%%%%%%%%%%%%%%%%%%%%%%%%%%%%%%%%%%%%%%%%%%%%%%%%%%%%%%%%%%%%%%%%%%

%%%%%%%%%%%%%%%%%%%%%%%%%%%%%%%%%%%%%%%%%%%%%%%%%%%%%%%%%%%%%%%%%%%%%%%%%%%%%%%%

%%%%%%%%%%%%%%%%%%%%%%%%%%%%%%%%%%%%%%%%%%%%%%%%%%%%%%%%%%%%%%%%%%%%%%%%%%%%%%%%
\newpage
\bibliographystyle{IEEEtran}
\balance
\bibliography{main}

\begin{thebibliography}{10}
\providecommand{\url}[1]{#1}
\csname url@samestyle\endcsname
\providecommand{\newblock}{\relax}
\providecommand{\bibinfo}[2]{#2}
\providecommand{\BIBentrySTDinterwordspacing}{\spaceskip=0pt\relax}
\providecommand{\BIBentryALTinterwordstretchfactor}{4}
\providecommand{\BIBentryALTinterwordspacing}{\spaceskip=\fontdimen2\font plus
\BIBentryALTinterwordstretchfactor\fontdimen3\font minus
  \fontdimen4\font\relax}
\providecommand{\BIBforeignlanguage}[2]{{%
\expandafter\ifx\csname l@#1\endcsname\relax
\typeout{** WARNING: IEEEtran.bst: No hyphenation pattern has been}%
\typeout{** loaded for the language `#1'. Using the pattern for}%
\typeout{** the default language instead.}%
\else
\language=\csname l@#1\endcsname
\fi
#2}}
\providecommand{\BIBdecl}{\relax}
\BIBdecl

\bibitem{c1}
E.~Tolstaya, R.~Mahjourian, C.~Downey, B.~Vadarajan, B.~Sapp, and D.~Anguelov,
  ``Identifying driver interactions via conditional behavior prediction,'' in
  \emph{2021 IEEE International Conference on Robotics and Automation
  (ICRA)}.\hskip 1em plus 0.5em minus 0.4em\relax IEEE, 2021, pp. 3473--3479.

\bibitem{c2}
Q.~Sun, X.~Huang, J.~Gu, B.~C. Williams, and H.~Zhao, ``M2i: From factored
  marginal trajectory prediction to interactive prediction,'' in
  \emph{Proceedings of the IEEE/CVF Conference on Computer Vision and Pattern
  Recognition}, 2022, pp. 6543--6552.

\bibitem{c3}
J.~L.~V. Espinoza, A.~Liniger, W.~Schwarting, D.~Rus, and L.~Van~Gool, ``Deep
  interactive motion prediction and planning: Playing games with motion
  prediction models,'' in \emph{Learning for Dynamics and Control
  Conference}.\hskip 1em plus 0.5em minus 0.4em\relax PMLR, 2022, pp.
  1006--1019.

\bibitem{c4}
X.~Tang, K.~Yang, H.~Wang, J.~Wu, Y.~Qin, W.~Yu, and D.~Cao,
  ``Prediction-uncertainty-aware decision-making for autonomous vehicles,''
  \emph{IEEE Transactions on Intelligent Vehicles}, vol.~7, no.~4, pp.
  849--862, 2022.

\bibitem{c5}
Z.~Huang, J.~Wu, and C.~Lv, ``Efficient deep reinforcement learning with
  imitative expert priors for autonomous driving,'' \emph{IEEE Transactions on
  Neural Networks and Learning Systems}, vol.~34, no.~10, pp. 7391--7403, 2022.

\bibitem{c6}
A.~Kendall, J.~Hawke, D.~Janz, P.~Mazur, D.~Reda, J.-M. Allen, V.-D. Lam,
  A.~Bewley, and A.~Shah, ``Learning to drive in a day,'' in \emph{2019
  international conference on robotics and automation (ICRA)}.\hskip 1em plus
  0.5em minus 0.4em\relax IEEE, 2019, pp. 8248--8254.

\bibitem{c7}
S.~Aradi, ``Survey of deep reinforcement learning for motion planning of
  autonomous vehicles,'' \emph{IEEE Transactions on Intelligent Transportation
  Systems}, vol.~23, no.~2, pp. 740--759, 2020.

\bibitem{c9}
X.~Li, X.~Ying, and M.~C. Chuah, ``Grip: Graph-based interaction-aware
  trajectory prediction,'' in \emph{2019 IEEE Intelligent Transportation
  Systems Conference (ITSC)}.\hskip 1em plus 0.5em minus 0.4em\relax IEEE,
  2019, pp. 3960--3966.

\bibitem{c10}
J.~Gao, C.~Sun, H.~Zhao, Y.~Shen, D.~Anguelov, C.~Li, and C.~Schmid,
  ``Vectornet: Encoding hd maps and agent dynamics from vectorized
  representation,'' in \emph{Proceedings of the IEEE/CVF conference on computer
  vision and pattern recognition}, 2020, pp. 11\,525--11\,533.

\bibitem{c11}
B.~Varadarajan, A.~Hefny, A.~Srivastava, K.~S. Refaat, N.~Nayakanti,
  A.~Cornman, K.~Chen, B.~Douillard, C.~P. Lam, D.~Anguelov \emph{et~al.},
  ``Multipath++: Efficient information fusion and trajectory aggregation for
  behavior prediction,'' in \emph{2022 International Conference on Robotics and
  Automation (ICRA)}.\hskip 1em plus 0.5em minus 0.4em\relax IEEE, 2022, pp.
  7814--7821.

\bibitem{c12}
Y.~Chai, B.~Sapp, M.~Bansal, and D.~Anguelov, ``Multipath: Multiple
  probabilistic anchor trajectory hypotheses for behavior prediction,''
  \emph{arXiv preprint arXiv:1910.05449}, 2019.

\bibitem{c13}
H.~Cui, V.~Radosavljevic, F.-C. Chou, T.-H. Lin, T.~Nguyen, T.-K. Huang,
  J.~Schneider, and N.~Djuric, ``Multimodal trajectory predictions for
  autonomous driving using deep convolutional networks,'' in \emph{2019
  international conference on robotics and automation (icra)}.\hskip 1em plus
  0.5em minus 0.4em\relax IEEE, 2019, pp. 2090--2096.

\bibitem{c14}
T.~N. Kipf and M.~Welling, ``Semi-supervised classification with graph
  convolutional networks,'' \emph{arXiv preprint arXiv:1609.02907}, 2016.

\bibitem{c15}
P.~Veli{\v{c}}kovi{\'c}, G.~Cucurull, A.~Casanova, A.~Romero, P.~Lio, and
  Y.~Bengio, ``Graph attention networks,'' \emph{arXiv preprint
  arXiv:1710.10903}, 2017.

\bibitem{c16}
J.~Gu, C.~Sun, and H.~Zhao, ``Densetnt: End-to-end trajectory prediction from
  dense goal sets,'' in \emph{Proceedings of the IEEE/CVF International
  Conference on Computer Vision}, 2021, pp. 15\,303--15\,312.

\bibitem{c17}
J.~Gao, C.~Sun, H.~Zhao, Y.~Shen, D.~Anguelov, C.~Li, and C.~Schmid,
  ``Vectornet: Encoding hd maps and agent dynamics from vectorized
  representation,'' in \emph{Proceedings of the IEEE/CVF conference on computer
  vision and pattern recognition}, 2020, pp. 11\,525--11\,533.

\bibitem{c18}
Z.~Huang, X.~Mo, and C.~Lv, ``Multi-modal motion prediction with
  transformer-based neural network for autonomous driving,'' in \emph{2022
  International Conference on Robotics and Automation (ICRA)}.\hskip 1em plus
  0.5em minus 0.4em\relax IEEE, 2022, pp. 2605--2611.

\bibitem{c19}
N.~Deo and M.~M. Trivedi, ``Convolutional social pooling for vehicle trajectory
  prediction,'' in \emph{Proceedings of the IEEE conference on computer vision
  and pattern recognition workshops}, 2018, pp. 1468--1476.

\bibitem{c20}
Y.~Wang, J.~Wang, J.~Jiang, S.~Xu, and J.~Wang, ``Sa-lstm: A trajectory
  prediction model for complex off-road multi-agent systems considering
  situation awareness based on risk field,'' \emph{IEEE Transactions on
  Vehicular Technology}, vol.~72, no.~11, pp. 14\,016--14\,027, 2023.

\bibitem{c21}
K.~Yang, B.~Li, W.~Shao, X.~Tang, X.~Liu, and H.~Wang, ``Prediction failure
  risk-aware decision-making for autonomous vehicles on signalized
  intersections,'' \emph{IEEE Transactions on Intelligent Transportation
  Systems}, 2023.

\bibitem{c22}
M.~Geisslinger, F.~Poszler, and M.~Lienkamp, ``An ethical trajectory planning
  algorithm for autonomous vehicles,'' \emph{Nature Machine Intelligence},
  vol.~5, no.~2, pp. 137--144, 2023.

\bibitem{c23}
T.~Wang, S.~Kim, J.~Wenxuan, E.~Xie, C.~Ge, J.~Chen, Z.~Li, and P.~Luo,
  ``Deepaccident: A motion and accident prediction benchmark for v2x autonomous
  driving,'' in \emph{Proceedings of the AAAI Conference on Artificial
  Intelligence}, vol.~38, no.~6, 2024, pp. 5599--5606.

\bibitem{c24}
A.~Alahi, K.~Goel, V.~Ramanathan, A.~Robicquet, L.~Fei-Fei, and S.~Savarese,
  ``Social lstm: Human trajectory prediction in crowded spaces,'' in
  \emph{Proceedings of the IEEE conference on computer vision and pattern
  recognition}, 2016, pp. 961--971.

\bibitem{c26}
A.~Gupta, J.~Johnson, L.~Fei-Fei, S.~Savarese, and A.~Alahi, ``Social gan:
  Socially acceptable trajectories with generative adversarial networks,'' in
  \emph{Proceedings of the IEEE conference on computer vision and pattern
  recognition}, 2018, pp. 2255--2264.

\bibitem{c27}
Z.~Huang, H.~Liu, J.~Wu, and C.~Lv, ``Differentiable integrated motion
  prediction and planning with learnable cost function for autonomous
  driving,'' \emph{IEEE transactions on neural networks and learning systems},
  2023.

\bibitem{c28}
X.~Chen, H.~Zhang, F.~Zhao, Y.~Cai, H.~Wang, and Q.~Ye, ``Vehicle trajectory
  prediction based on intention-aware non-autoregressive transformer with
  multi-attention learning for internet of vehicles,'' \emph{IEEE Transactions
  on Instrumentation and Measurement}, vol.~71, pp. 1--12, 2022.

\bibitem{c29}
X.~Tang, K.~Yang, H.~Wang, W.~Yu, X.~Yang, T.~Liu, and J.~Li, ``Driving
  environment uncertainty-aware motion planning for autonomous vehicles,''
  \emph{Chinese Journal of Mechanical Engineering}, vol.~35, no.~1, p. 120,
  2022.

\bibitem{c30}
M.~Koschi, C.~Pek, M.~Beikirch, and M.~Althoff, ``Set-based prediction of
  pedestrians in urban environments considering formalized traffic rules,'' in
  \emph{2018 21st international conference on intelligent transportation
  systems (ITSC)}.\hskip 1em plus 0.5em minus 0.4em\relax IEEE, 2018, pp.
  2704--2711.

\bibitem{c31}
M.~Althoff and S.~Magdici, ``Set-based prediction of traffic participants on
  arbitrary road networks,'' \emph{IEEE Transactions on Intelligent Vehicles},
  vol.~1, no.~2, pp. 187--202, 2016.

\bibitem{c32}
G.~Alinezhad~Noghre, V.~Katariya, A.~Danesh~Pazho, C.~Neff, and H.~Tabkhi,
  ``Pishgu: Universal path prediction network architecture for real-time
  cyber-physical edge systems,'' in \emph{Proceedings of the ACM/IEEE 14th
  International Conference on Cyber-Physical Systems (with CPS-IoT Week 2023)},
  2023, pp. 88--97.

\bibitem{c33}
X.~Wang, K.~Tang, X.~Dai, J.~Xu, J.~Xi, R.~Ai, Y.~Wang, W.~Gu, and C.~Sun,
  ``Safety-balanced driving-style aware trajectory planning in intersection
  scenarios with uncertain environment,'' \emph{IEEE Transactions on
  Intelligent Vehicles}, vol.~8, no.~4, pp. 2888--2898, 2023.

\bibitem{c34}
X.~Tang, K.~Yang, H.~Wang, W.~Yu, X.~Yang, T.~Liu, and J.~Li, ``Driving
  environment uncertainty-aware motion planning for autonomous vehicles,''
  \emph{Chinese Journal of Mechanical Engineering}, vol.~35, no.~1, p. 120,
  2022.

\bibitem{wang1}
J.~Wang, W.~Chi, C.~Li, C.~Wang, and M.~Q.-H. Meng, ``Neural rrt*:
  Learning-based optimal path planning,'' \emph{IEEE Transactions on Automation
  Science and Engineering}, vol.~17, no.~4, pp. 1748--1758, 2020.

\bibitem{wang2}
J.~Wang, W.~Chen, X.~Xiao, Y.~Xu, C.~Li, X.~Jia, and M.~Q.-H. Meng, ``A survey
  of the development of biomimetic intelligence and robotics,''
  \emph{Biomimetic Intelligence and Robotics}, vol.~1, p. 100001, 2021.

\end{thebibliography}

\end{document}